\documentclass{article}

\usepackage{amssymb}
\usepackage{graphicx}

\usepackage{times}
\usepackage{helvet}
\usepackage{courier}
\newtheorem{theorem}{Theorem}
\newtheorem{assumption}{Assumption}
\newtheorem{proof}{Proof}

\newtheorem{lemma}[theorem]{Lemma}
\usepackage{algorithm}
\usepackage{algorithmic}
\usepackage{amsmath}
\usepackage{subcaption}
\usepackage{booktabs}
\usepackage{epsfig}
\usepackage{epstopdf}
\usepackage{url}
\usepackage{multirow}

\title{ Decoupled Asynchronous Proximal Stochastic Gradient Descent with Variance Reduction}

\begin{document}

\author{Zhouyuan Huo\\
zhouyuan.huo@mavs.uta.edu \\
\and
Bin Gu\\
jsgubin@gmail.com
\and
Heng Huang\\
heng@uta.edu
}

\maketitle

\begin{abstract}
In the era of big data, optimizing large scale machine learning problems becomes a challenging task and draws significant attention. Asynchronous optimization algorithms come out as a promising solution. Recently, decoupled asynchronous proximal stochastic gradient descent (DAP-SGD) is proposed to minimize a composite function. It is claimed to be able to offload the computation bottleneck from server to workers by allowing workers to evaluate the proximal operators, therefore, server just need to do element-wise operations. However, it still suffers from slow convergence rate because of the variance of stochastic gradient is not zero. In this paper, we propose a faster method, decoupled asynchronous proximal stochastic variance reduced gradient descent method (DAP-SVRG). We prove that our method has linear convergence for strongly convex problem. %Large-scale experiments are also conducted in this paper, and results demonstrate our theoretical analysis.
\end{abstract}

\section{Introduction}
In this paper, we consider the problem of minimizing the following composite functions:
\begin{eqnarray}
\min_{x \in \mathbb{R}^d} P(x)  = f(x) + h(x) \nonumber \\ 
f(x) = \frac{1}{n} \sum\limits_{i=1}^n f_i(x)
\end{eqnarray}
where $f(x)$ is the average of a number of convex functions, and $h(x)$ is a regularization term which is usually non-differentiable. There are many machine learning problems in this form. In ridge regression, $f_i(x) = (x^Ta_i - b_i)^2$ and $h(x) = \frac{\lambda}{2}\|x\|_2^2$, where $(a_i,b_i)$ is the data sample we collect. If regularization $h(x) = \lambda \|x\|_1$, it becomes Lasso problem, and if $h(x) = \lambda_1 \|x\|_1 + \frac{\lambda_2}{2} \|x\|^2_2$, it becomes elastic net problem.  

Stochastic methods \cite{rosasco2014convergence,konevcny2014ms2gd,xiao2014proximal,nitanda2014stochastic} are advantageous for this kind of problem when $n$ is very large, in each iteration, only a single or a mini-batch samples are used for update. Because the variance of approximate gradient is not zero, a decreasing learning rate has to be used to obtain optimal solution, and it also leads to slow convergence rate. Proximal stochastic gradient descent (Prox-SGD) \cite{rosasco2014convergence} is the most popular method to solve regularized empirical loss minimization problem, and it is proved to have a convergence rate of $O(1/T)$. In \cite{zhong2014accelerated}, authors propose an accelerated stochastic gradient method for composite regularization by using Nesterov's acceleration technique. It converges with rate $O(1/T^2)$. Recently, variance reduction technique \cite{johnson2013accelerating,defazio2014saga} makes linear convergence rate real. In \cite{xiao2014proximal,nitanda2014stochastic} variance reduction technique is used and proximal stochastic gradient descent method with variance reduction (Prox-SVRG) is proved to converge linearly.  When $h(x) =  \frac{\lambda}{2}\|x\|_2^2$, stochastic dual coordinate ascent (SDCA) performs better in some regimes especially SVM problem \cite{hsieh2008dual,shalev2013stochastic}, when $f_i(x) = \max (0, 1- b_ix^Ta_i)$. 

With the proliferation of information, the size of data grows larger, and the optimization of large-scale machine learning model becomes a challenging problem. To handle large-scale data, stochastic gradient descent and its variants are proposed for its low computation complexity per iteration \cite{bottou2010large}. However, when dataset can not be stored and processed in a single machine, we have to use distributed optimization method. There are many works on distributed machine learning systems like Parameter Server \cite{li2014communication} or GraphLab \cite{low2012distributed}. Traditional serial optimization methods are extended to distributed version for large-scale problem, such as stochastic gradient descent and its variants(SGD) \cite{lian2015asynchronous,zhang2015fast,langford2009slow,zhao2016fast,recht2011hogwild,huo2016asynchronous}, stochastic coordinate descent (SCD) \cite{liu2014asynchronous}, alternating direction method of multipliers (ADMM) \cite{zhang2014asynchronous}, stochastic dual coordinate ascent (SDCA) \cite{yang2013trading,jaggi2014communication,takac2015distributed}. 
There are mainly two architectures in distributed system: one is shared-memory architecture, and the other is distributed-memory architecture. Distributed-memory architecture is about the distributed system with multiple machines, in this paper, we consider distributed-memory architecture with one server and multiple workers. All the communication happens between server and each worker. There are mainly two types of communication, synchronous communication and asynchronous communication. In synchronous method, server has to wait and get information from all workers before it goes to the next step. It is known that the synchronous method suffers from straggler problem. To alleviate the communication overhead, asynchronous methods are proposed. In these methods, server can update parameters with stale information. Although the stale gradient information introduces noise in the process, it does not necessarily affect the convergence property of asynchronous method. There are also some research to alleviate this noise by using staleness variable in the update rule \cite{zhang2015staleness,odena2016faster}. Throughout this paper, we only consider asynchronous communication between each worker and server. 

In \cite{li2016make}, they claim that running proximal step in server is time-consuming and the propose a decoupled asynchronous proximal stochastic gradient method (DAP-SGD), which off-loads the proximal step to workers, and the server just needs to do element-wise computation. Experimental results show that DAP-SGD converges faster than traditional asynchronous proximal stochastic gradient method (TAP-SGD). Although its accelerate in running time, it just has convergence rate $O(1/\sqrt{T})$ on strongly convex and smooth problems when learning rate is constant. In this paper, we propose a decoupled asynchronous  proximal stochastic gradient with variance reduction (DAP-SVRG), and we prove that it has linear convergence for strongly convex problem.

\section{Traditional Asynchronous Proximal Stochastic Gradient Descent with Variance Reduction (TAP-SVRG)}

The pseudocode of TAP-SVRG for worker and server are provided in Algorithm \ref{alg1tap} and Algorithm \ref{alg2tap} respectively. In each worker, it computes variance reduced gradient $v_t$ of random sample:
\begin{eqnarray}
v_t = \nabla f_i(x_{d(t)}) - \nabla f_i(\tilde x) + \nabla f(\tilde x)
\end{eqnarray}
 and send it to server, where $x_{d(t)}$ denotes stale parameter used for update in iteration $t$, $\tilde x$ is a snapshot of parameter after every $T$ iteration.  In TAP-SVRG, proximal mapping happens in server side:
 \begin{eqnarray}
x_{s,t+1} = \text{Prox}_{\eta, h}(x_{s,t} - \eta v_t)
 \end{eqnarray}
 where proximal operator $\text{Prox}_{\eta, h} \arg\min_y \frac{1}{2\eta}\|y-x\|_2^2 + h(y)$.

\begin{algorithm}                      % enter the algorithm environment
	\caption{TAP-SVRG (Worker $k$)}         % give the algorithm a caption
	\label{alg1tap}                           % and a label for \ref{} commands later in the document
	\begin{algorithmic}
		\IF{$flag$ is True}
		\STATE Pull parameter $\tilde x$ from server.  \\
		\STATE Evaluate and send gradient $\nabla f^k(\tilde x) \leftarrow \sum\limits_{i=1}^{n_k} \nabla f_i(\tilde x)$ to server. \\
		\STATE Receive  $\nabla f(\tilde x)$ from server. \\
		\ELSE 
		\STATE Receive parameter $x_{d(t)}$ from server.  \\
		\STATE Uniformly sample $i$ from $[1,...,n_k ]$.\\
		\STATE Evaluate the variance reduced gradient of sample $i$ over parameter $x_{d(t)}$. 
		\begin{equation}v_t \leftarrow \nabla f_i(x_{d(t)}) - \nabla f_i(\tilde x) + \nabla f(\tilde x) \end{equation}
		%\STATE Evaluate the proximal operator $x_{d(t)}' = \text{Prox}_{\eta, h}(x_{d(t)} - \eta \nabla v_i)$. \\
		%\STATE Send update information $\Delta  \leftarrow x_{d(t)}' - x_{d(t)}$ to master. \\
		\STATE Send update information $v_t$ to server.
		\ENDIF
	\end{algorithmic}
\end{algorithm}

\begin{algorithm}                      % enter the algorithm environment
	\caption{TAP-SVRG (Server)}         % give the algorithm a caption
	\label{alg2tap}                           % and a label for \ref{} commands later in the document
	\begin{algorithmic}
		\FOR{$s = 0 \text{ to } S-1$}
		\STATE $ flag = \text{True}$. \\
		\STATE Broadcast $flag$ and  $\tilde{x}_s$ to $K$ workers. \\
		\STATE Receive $\nabla f^k(\tilde x)$ from workers. \\
		\STATE Compute full gradient $\nabla f(\tilde x_s) \leftarrow \frac{1}{n} \sum\limits_{k=1}^K \nabla f^k(\tilde x_s)$. \\
		\STATE $ flag = \text{False}$. \\
		\STATE Broadcast $flag$  to $K$ workers. \\
		
		\FOR{$t = 0 \text{ to } T-1$}
		%\STATE  Receive $\Delta_{d(t)} = x'_{d(t)} - x_{d(t)}$ from one worker. \\
		%\STATE  Update parameter with $x_{s,t+1} = x_{s,t} + \Delta_{d(t)}$. \\
		\STATE Receive $v_t$ from worker. \\
	  \STATE Evaluate the proximal operator $x_{s,t+1} \leftarrow \text{Prox}_{\eta, h}(x_{s,t} - \eta v_t)$. \\
		\ENDFOR
		\STATE $x_{s+1,0} = x_{s,T}$\\
		\STATE $\tilde{x}_{s+1} =x_{s,T}$  \\
		\ENDFOR
		
	\end{algorithmic}
\end{algorithm}

\section{Decoupled Asynchronous Proximal Stochastic Gradient Descent with Variance Reduction (DAP-SVRG)}

In \cite{li2016make}, authors find that if proximal mapping is time-consuming, the update procedure in server is the computational bottleneck. Because workers have to wait until the mapping in the server is done. To avoid this problem, as per \cite{li2016make}, we make workers do proximal mapping step, and server is just responsible for element-wise addition operations. Pseudocode of DAP-SVRG for worker nd server are summarized in Algorithm \ref{alg1dap} and Algorithm \ref{alg2dap} respectively.  In each iteration, workers receive stale parameter $x_{d(t)}$ from server, and computes $v_t$, then do proximal mapping: 
\begin{eqnarray}
	x_{d(t)}' = \text{Prox}_{\eta, h}(x_{d(t)} - \eta v_t)
\end{eqnarray} 
Finally, workers send $\Delta_{d(t)} = x'_{d(t)} - x_{d(t)}$ to server. Therefore, in the server, we just need to do element-wise addition operations, and it is easy to parallelize: 
\begin{eqnarray}
x_{s,t+1} = x_{s,t} + \Delta_{d(t)}
\end{eqnarray}

\begin{algorithm}                      % enter the algorithm environment
\caption{DAP-SVRG (Worker $k$)}         % give the algorithm a caption
\label{alg1dap}                           % and a label for \ref{} commands later in the document
\begin{algorithmic}
\IF{$flag$ is True}
  \STATE Pull parameter $\tilde x$ from server.  \\
		\STATE Evaluate and send gradient $\nabla f^k(\tilde x) \leftarrow \sum\limits_{i=1}^{n_k} \nabla f_i(\tilde x)$ to server. \\
  \STATE Receive  $\nabla f(\tilde x)$ from server. \\
\ELSE 
  \STATE Receive parameter $x_{d(t)}$ from server.  \\
  \STATE Uniformly sample $i$ from $[1,...,n_k ]$.\\
  \STATE Evaluate the variance reduced gradient of sample $i$ over parameter $x_{d(t)}$. 
  \begin{equation}v_t \leftarrow \nabla f_i(x_{d(t)}) - \nabla f_i(\tilde x) + \nabla f(\tilde x) \end{equation}
  \STATE Evaluate the proximal operator $x_{d(t)}' \leftarrow \text{Prox}_{\eta, h}(x_{d(t)} - \eta v_t)$. \\
  \STATE Send update information $\Delta  \leftarrow x_{d(t)}' - x_{d(t)}$ to server. \\
  \ENDIF
\end{algorithmic}
\end{algorithm}

\begin{algorithm}                      % enter the algorithm environment
\caption{DAP-SVRG (Server)}         % give the algorithm a caption
\label{alg2dap}                           % and a label for \ref{} commands later in the document
\begin{algorithmic}
\FOR{$s = 0 \text{ to } S-1$}
  \STATE $ flag = \text{True}$. \\
  \STATE Broadcast $flag$ and  $\tilde{x}_s$ to $K$ workers. \\
  \STATE Receive $\nabla f^k(\tilde x)$ from workers. \\
  \STATE Compute full gradient $\nabla f(\tilde x_s) \leftarrow \frac{1}{n} \sum\limits_{k=1}^K \nabla f^k(\tilde x_s)$. \\
    \STATE $ flag = \text{False}$. \\
  \STATE Broadcast $flag$  to $K$ workers. \\

  \FOR{$t = 0 \text{ to } T-1$}
    \STATE  Receive $\Delta_{d(t)} = x'_{d(t)} - x_{d(t)}$ from one worker. \\
    \STATE  Update parameter with $x_{s,t+1} = x_{s,t} + \Delta_{d(t)}$. \\
  \ENDFOR
  \STATE $x_{s+1,0} = x_{s,T}$\\
  \STATE $\tilde{x}_{s+1} =x_{s,T}$  \\
\ENDFOR

\end{algorithmic}
\end{algorithm}

\section{Convergence Analysis}

In this section, we analyze the convergence property of our DAP-SVRG, and prove that it has linear convergence rate. Initially, we suppose that all the following assumptions are true, they are all common for theoretical analysis of asynchronous stochastic method. 

\begin{assumption}
f(x) is $u$ strongly convex. 
\begin{eqnarray}
f(x) \geq f(y) + \left<\nabla f(y), x-y \right> + \frac{\mu}{2}\|x-y\|_2^2
\end{eqnarray}
\end{assumption}

\begin{assumption}
f(x) is $L$ smooth. 
\begin{eqnarray}
f(x) \leq f(y) + \left< \nabla f(y), x-y \right> + \frac{L}{2}\|x-y \|_2^2
\end{eqnarray}
\end{assumption}

\begin{assumption}
h(x) is convex.
\begin{eqnarray}
h(x) \geq h(y) + \left< \partial h(y), x-y \right>,  \forall x,y
\end{eqnarray}
\end{assumption}

\begin{assumption}
The maximum time delay of  parameter $x$ in each worker is upper bounded by $\tau$.
\end{assumption}

If all these assumptions satisfy, at first, we bound the variance of stochastic gradient, and it goes to zero when $t$ goes to infinity.
\begin{lemma}
\label{lem1}
We define $x^*$ to be  the optimal solution of problem P(x), learning rate $\eta$, and $u_t$ is: 
\begin{eqnarray}
u_{t} = \nabla f_i(x_t) - \nabla f_i(\tilde{x}) + \nabla f(\tilde{x})
\end{eqnarray}
then, we know that:
\begin{eqnarray}
\sum\limits_{t=sm}^{sm+m-1} \mathbb{E}\|v_t - \nabla f(x_{d(t)}) \|^2_2 &\leq &  \frac{8L^2\tau^2 \eta}{1- 8L^2 \tau^2 \eta^2} \sum\limits_{t=sm}^{sm+m-1} \mathbb{E} \left[ P(x_t) - P(x_t') \right] \nonumber \\
&&+ \frac{2}{1- 8L^2 \tau^2 \eta^2} \sum\limits_{t=sm}^{sm+m-1} \mathbb{E} \|u_t - \nabla f(x_t)\|_2^2
\end{eqnarray}
\end{lemma}

Following Lemma \ref{lem1}, we are able to bound the variance of stochastic gradient, if $t$ goes to infinity, $P(x_t) - P(x_t')$ is close to zero, and $\mathbb{E}\|u_t - \nabla f(x_t)\|^2_2$ goes to zero too.  Then we have the convergence rate of DAP-SVRG.
\begin{theorem}
\label{thm1}
Define $x^*$ is the optimal solution, and learning rate $\eta$  and $\varepsilon$ are small positive constants, then we have:
\begin{eqnarray}
\mathbb{E} \left[ P(\tilde{x}_{s+1}) - P(x^*) \right]  \leq \rho \mathbb{E} \left[ P(\tilde{x}_{s}) - P(x^*) \right]
\end{eqnarray}
where 
\begin{eqnarray}
\rho = \frac{\frac{2}{\mu} + \frac{8L(6\eta^2 + 4 \eta^2 \tau^2)m}{1-8L^2\eta^2\tau^2} }{\left( 2\eta - \frac{8L^2 \eta^2 \tau \varepsilon (6\eta^2 + 4 \eta^2 \tau^2)}{1-8L^2\eta^2\tau^2} -  (6\eta + 4\eta \tau^2)\varepsilon - \frac{8L(6\eta^2 + 4 \eta^2 \tau^2)}{1-8L^2\eta^2\tau^2} \right)m}
\end{eqnarray}
\end{theorem}
Therefore, we know that DAP-SVRG has linear convergence rate when assumptions satisfy. 

\section{Experiments}
In this section, we present experimental results and evaluate the empirical performance of the proposed optimization algorithm. Our algorithm is implemented in C++, and the point to point communication is handled by openMPI\footnote{https://www.open-mpi.org/}. We use Armadillo library\footnote{http://arma.sourceforge.net/} for efficient matrix computation, such as SVD. We run our experiments on Amazon Web Services, and each node is a t2.medium instance which has 2 virtual CPUs. 

We consider the following problem with synthetic dataset:
\begin{eqnarray}
	\min_{X\in \mathbb{R}^{d_1 \times d_2}} \frac{1}{n} \sum\limits_{i=1}^n \|X^Ta_i - b_i\|_2^2 + \frac{\lambda_1}{2} \|X\|_F^2 +  \lambda_2\|X\|_*  
\end{eqnarray}
where $f(X) =  \frac{1}{n} \sum\limits_{i=1}^n \|X^Ta_i - b_i\|_2^2 + \frac{\lambda_1}{2} \|X\|_F^2$ and  $h(X)$ is nuclear norm $h(X) =   \lambda_2\|X\|_*$ which is very time-consuming to do its proximal mapping. 
In the experiment, we generate a low-rank matrix $X \in \mathbb{R}^{d_1 \times d_2}$, where $d_1 = 100$, $d_2=50$, and its rank is $10$. Then, we generate a matrix $A \in \mathbb{R}^{n \times d_1}$ randomly, where $n=10,000$. $B \in \mathbb{R}^{n \times d_2}$ is obtained through $B = AX$.  We set $\lambda_1 = 10^{-3}$ and $\lambda_2=10^{-3}$ respectively.      

\subsection{Faster Speedup}
\begin{figure}[t]
	\centering
	\begin{subfigure}[b]{0.45\textwidth}
		\centering
		\includegraphics[width=2.3in]{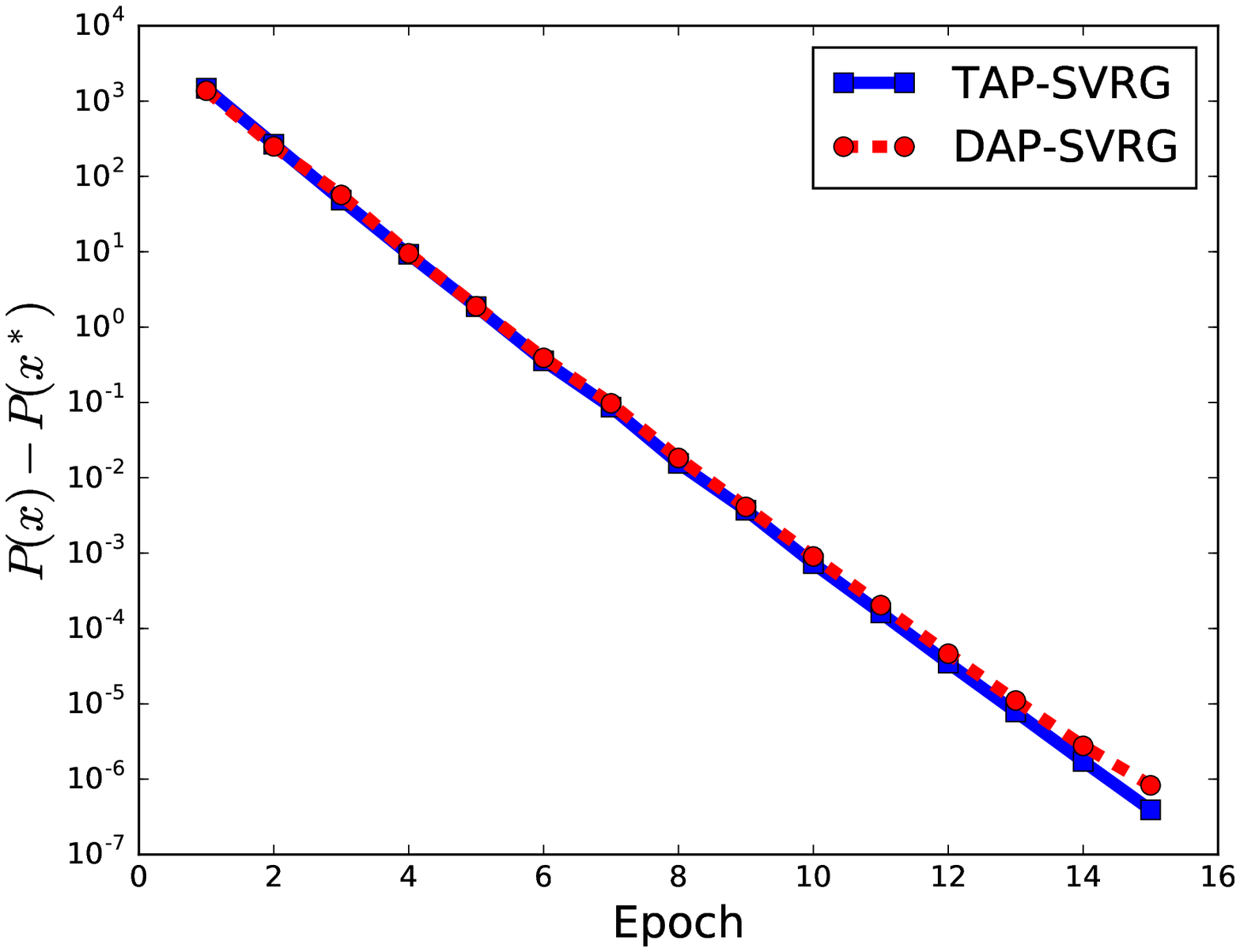}
		\caption{$P(x) - P(x^*)$  vs Epoch}
		\label{obj_epoch}
	\end{subfigure}
	\begin{subfigure}[b]{0.45\textwidth}
		\centering
		\includegraphics[width=2.3in]{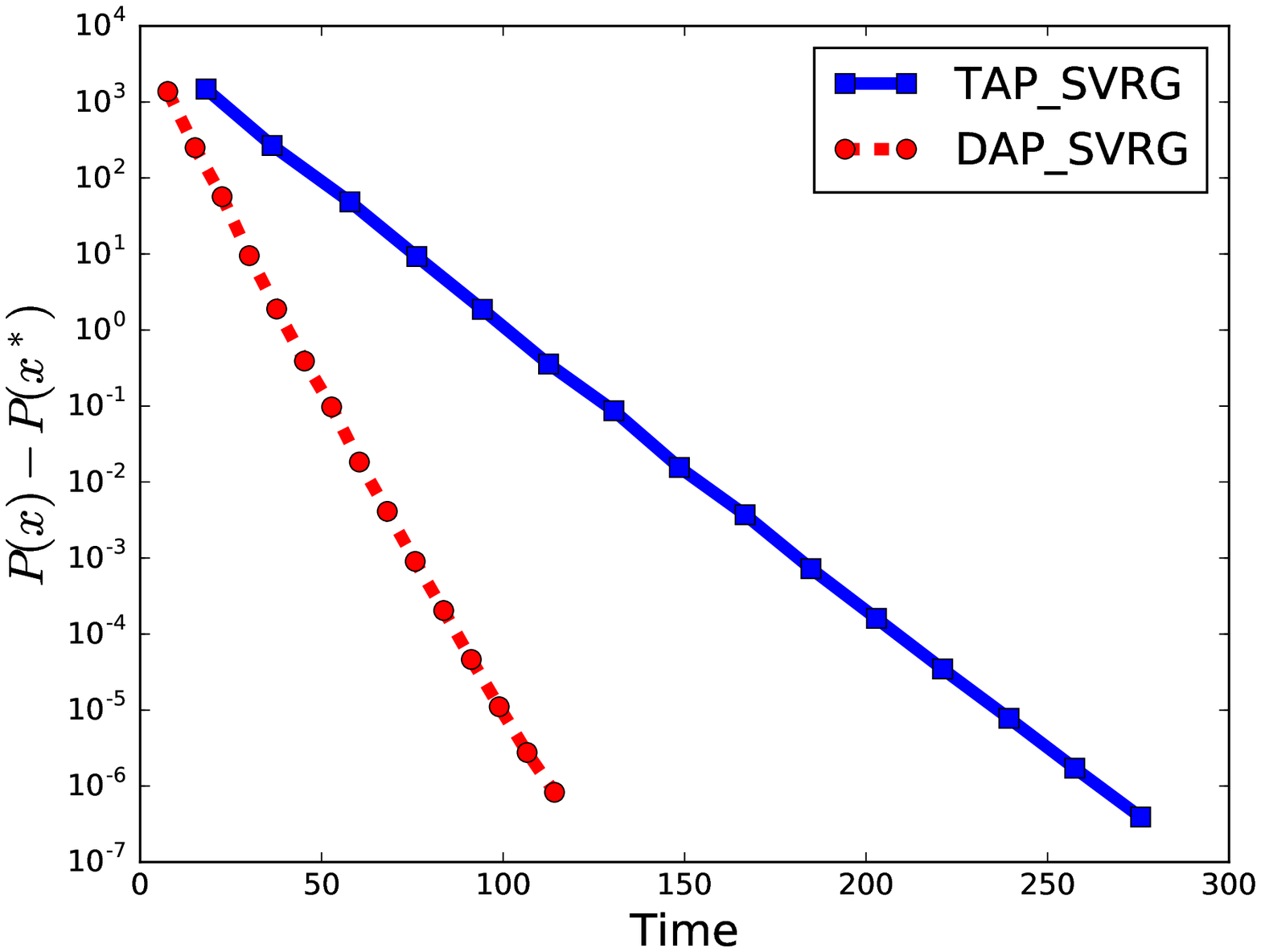}
		\caption{$P(x) - P(x^*)$  vs Time}
		\label{obj_time}
	\end{subfigure}
		\begin{subfigure}[b]{0.45\textwidth}
			\centering
			\includegraphics[width=2.3in]{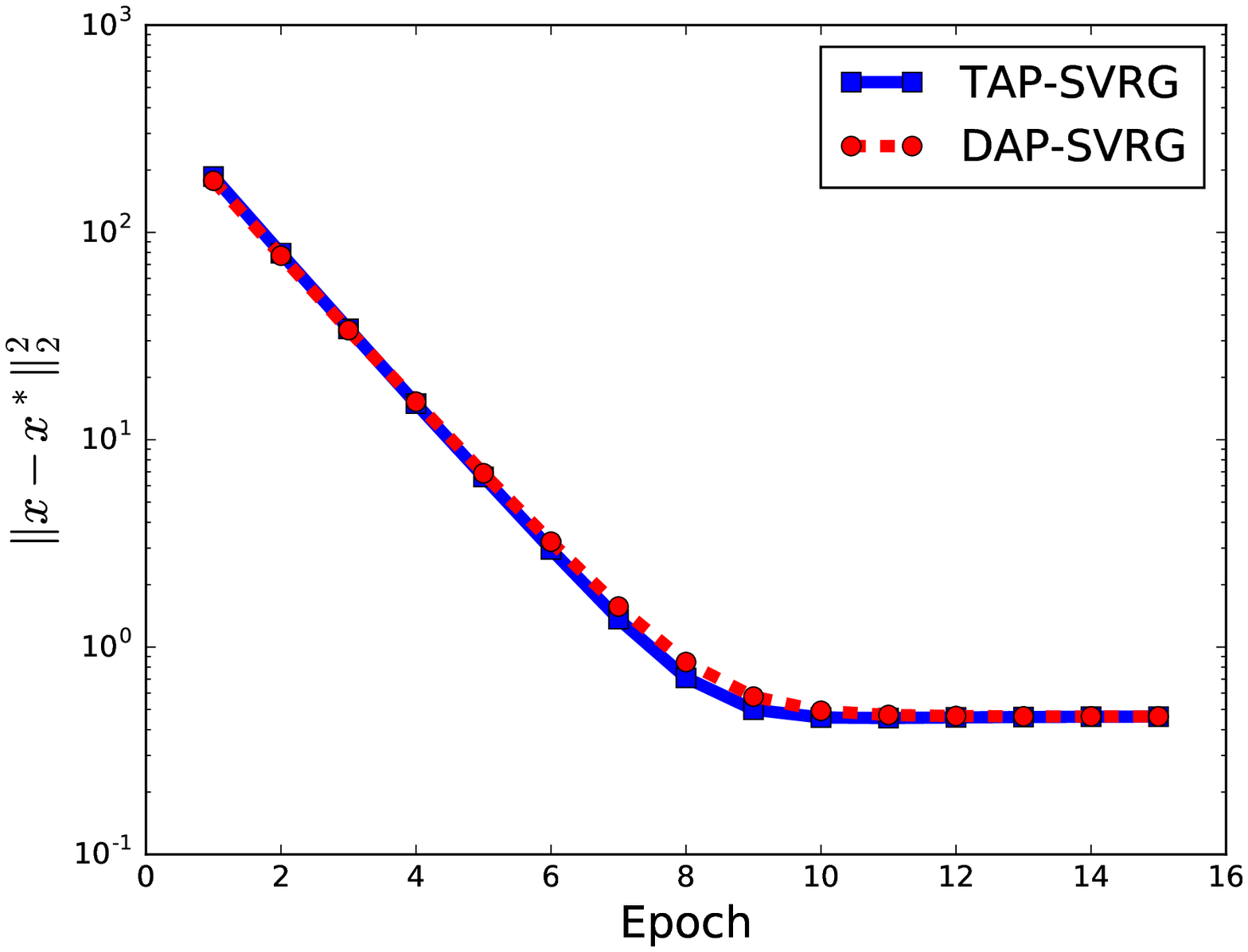}
			\caption{$\|x-x^*\|_2^2$ vs Epoch}
			\label{w_epoch}
		\end{subfigure}
		\begin{subfigure}[b]{0.45\textwidth}
			\centering
			\includegraphics[width=2.3in]{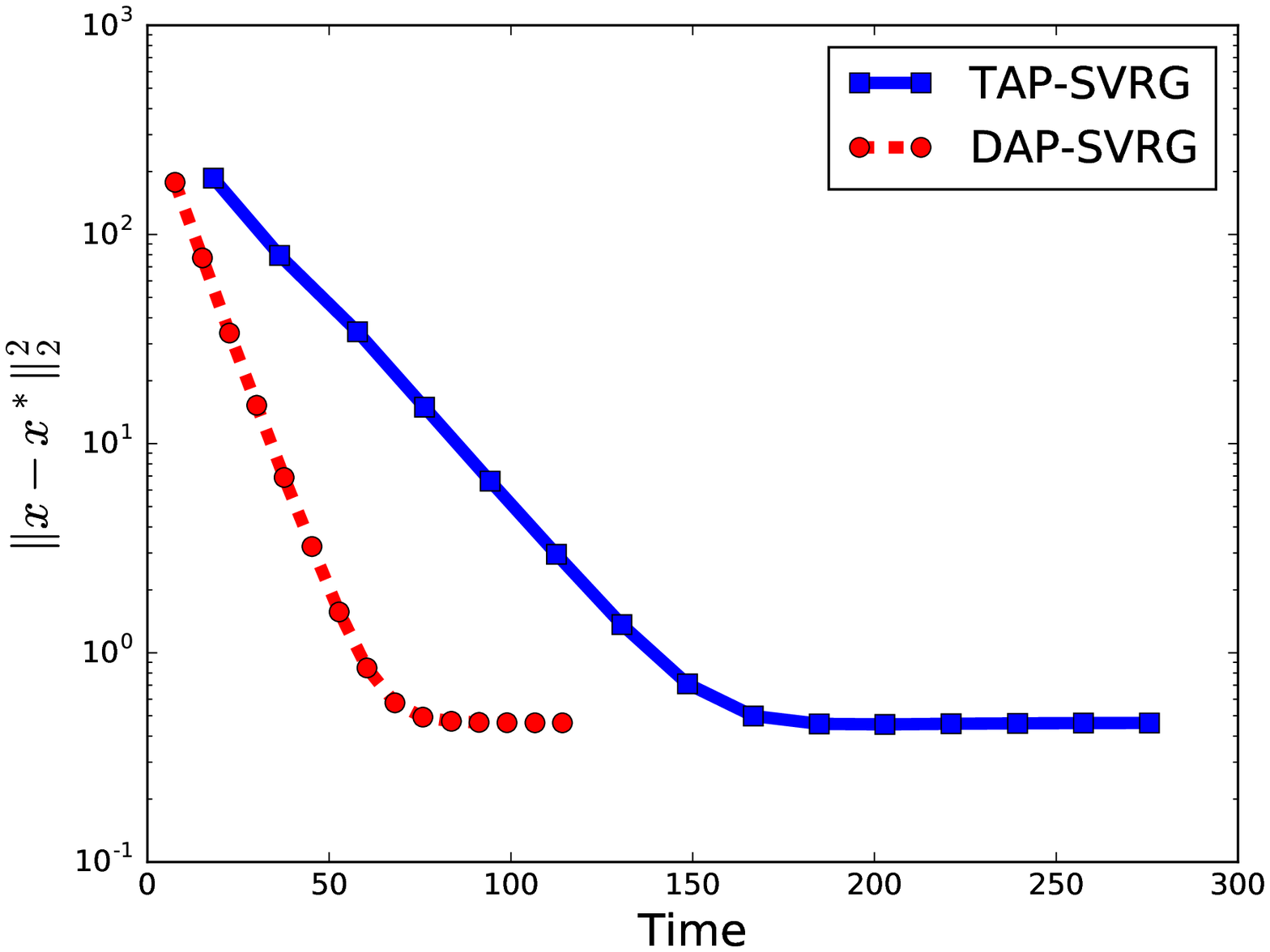}
			\caption{$\|x-x^*\|_2^2$ vs Time}
			\label{w_time}
		\end{subfigure}
	\caption{Performance of TAP-SVRG and DAP-SVRG for nuclear regularized objective. We evaluate the convergence of sub-optimality $P(x) - P(x^*)$ and $\|x - x^*\|_2^2$  in terms of both epoch and time.} 
	\label{multiworkers}
\end{figure}

In this section, we will verify that our DAP-SVRG enjoys better speedup than TAP-SVRG when we increase the number of workers.  We run DAP-SVRG and TAP-SVRG on a distributed system with $1$ server and $10$  workers. Experiment results are shown in Figure \ref{multiworkers}.  In Figure \ref{obj_epoch} and Figure \ref{w_epoch}, we know that our decoupled method does not affect convergence of $P(x) - P(x^*)$ and $\|x-x^*\|_2^2$ in terms of epoch or iterations. Figure \ref{obj_time} and Figure \ref{w_time} present that DAP-SVRG converges faster than TAP-SVRG in terms of time. In TAP-SVRG method, time-consuming nuclear proximal mapping is process in the server, thus information from workers hang in the air when server is busy. In our method, server just need to process additive operations, so it can handle more information from workers.

\begin{figure}
	\centering
	\includegraphics[width=3.3in]{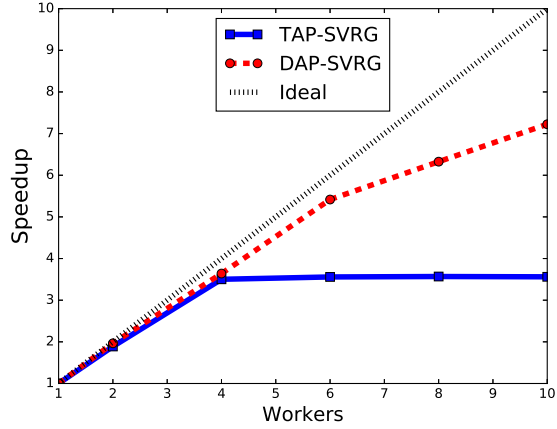}
	\caption{Time speedup when we increase the number of workers. The black dashed line denotes ideal speedup.}
	\label{fig_speedup}
\end{figure}

 We  also run TAP-SVRG and DAP-SVRG on distributed system and increase the number of workers from $1$ to $10$. Results  in Figure \ref{fig_speedup} show that our DAP-SVRG enjoys near linear speedup. For TAP-SVRG method, its speedup is limited due to proximal mapping in the server. It is obvious that when the number of workers is greater than $4$, increasing workers does not help anymore.
 
\subsection{Faster Convergence}
\begin{figure}[t]
	\centering
	\begin{subfigure}[b]{0.45\textwidth}
		\centering
		\includegraphics[width=2.3in]{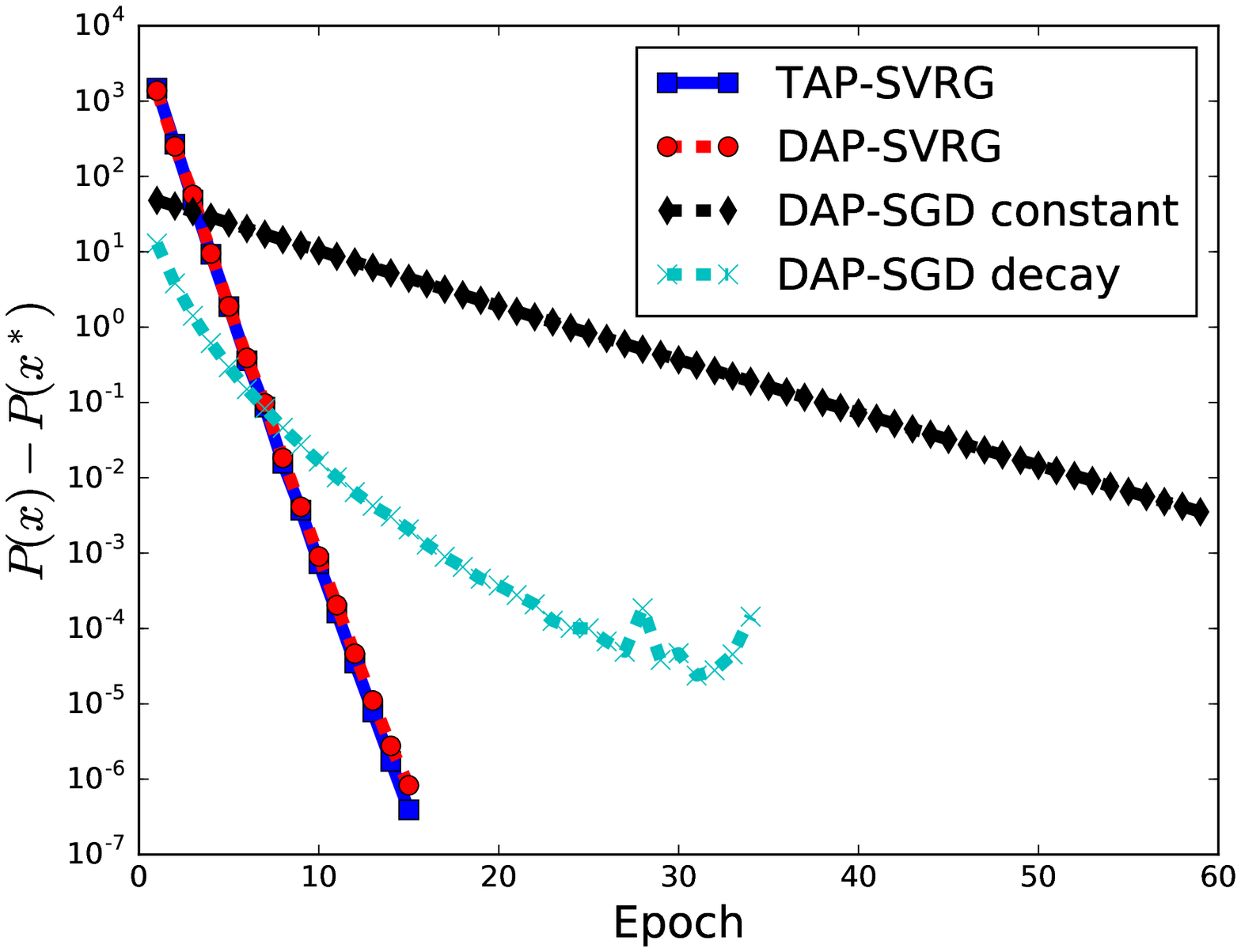}
		\caption{$P(x) - P(x^*)$  vs Epoch}
		\label{comp_obj_epoch}
	\end{subfigure}
	\begin{subfigure}[b]{0.45\textwidth}
		\centering
		\includegraphics[width=2.3in]{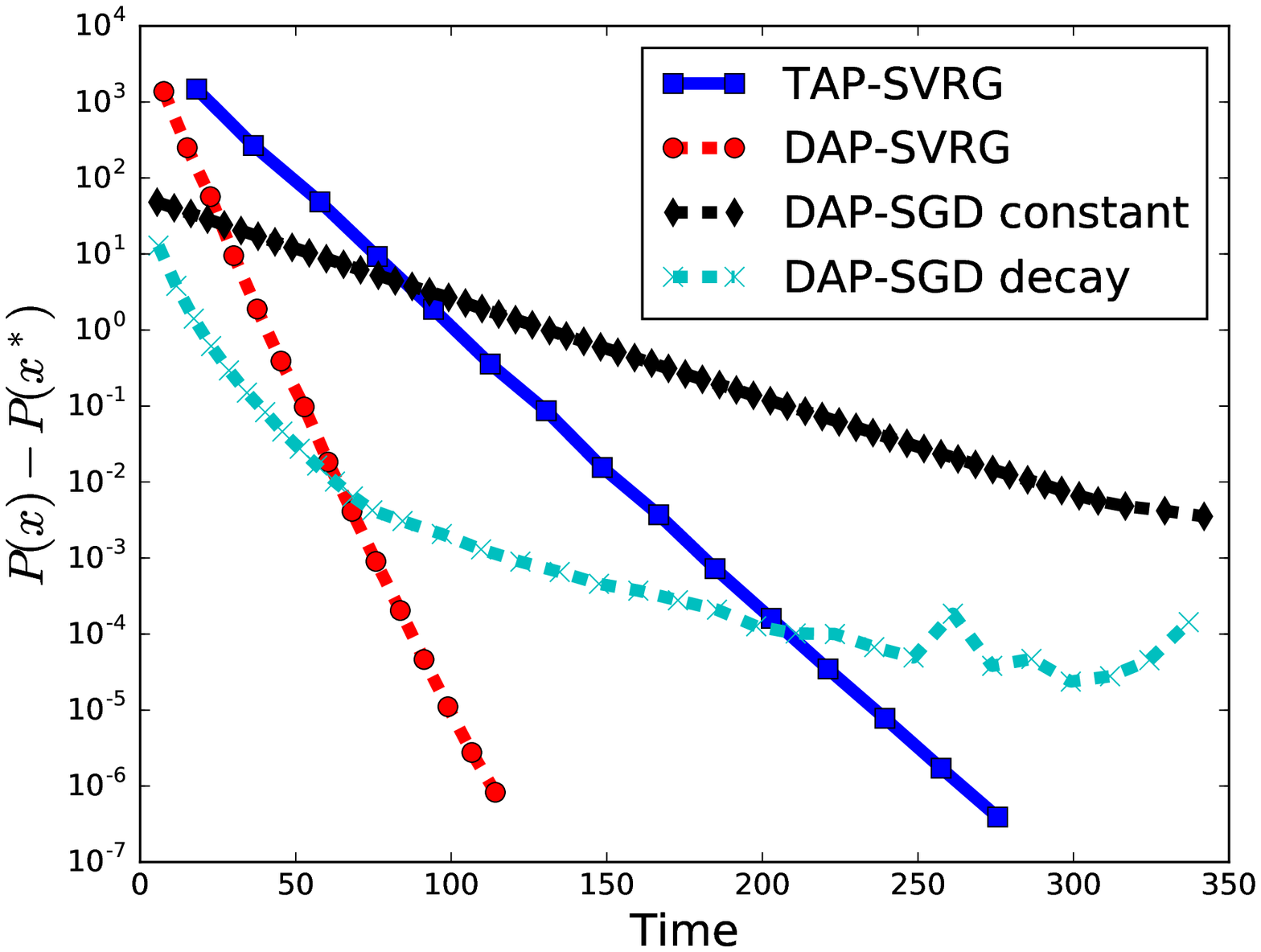}
		\caption{$P(x) - P(x^*)$  vs Time}
		\label{comp_obj_time}
	\end{subfigure}
	\begin{subfigure}[b]{0.45\textwidth}
		\centering
		\includegraphics[width=2.3in]{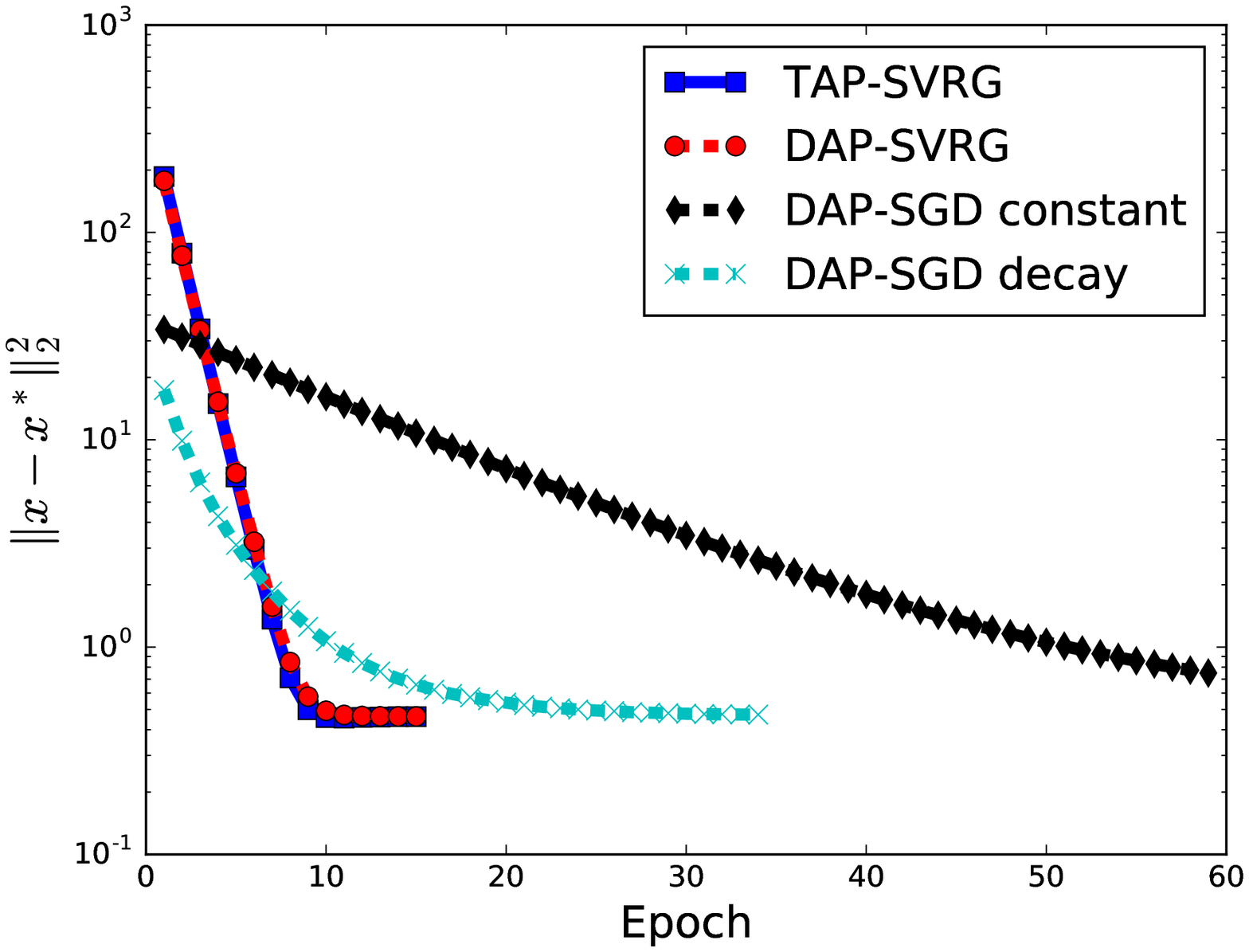}
		\caption{$\|x-x^*\|_2^2$ vs Epoch}
		\label{comp_w_epoch}
	\end{subfigure}
	\begin{subfigure}[b]{0.45\textwidth}
		\centering
		\includegraphics[width=2.3in]{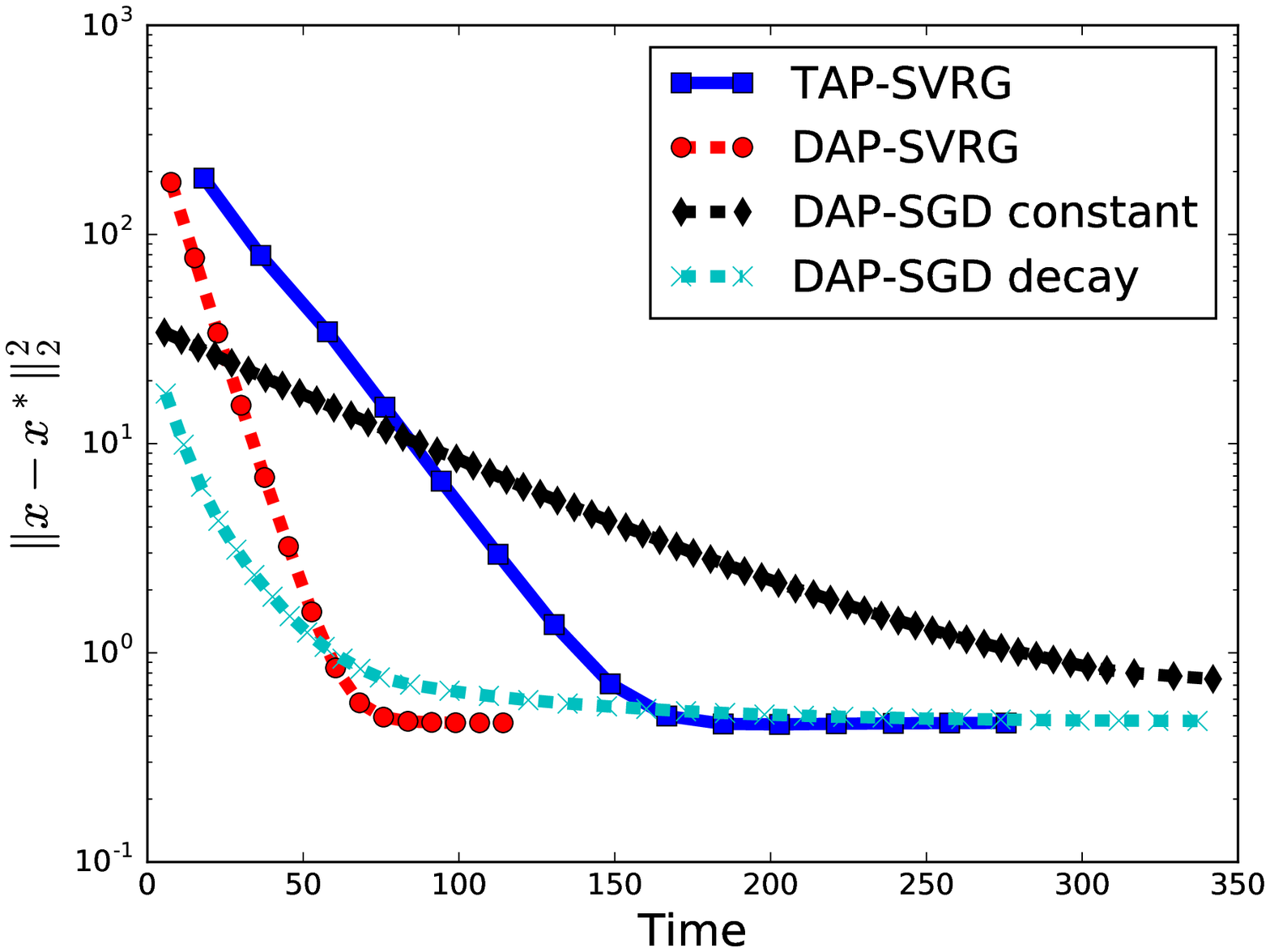}
		\caption{$\|x-x^*\|_2^2$ vs Time}
		\label{comp_w_time}
	\end{subfigure}
	\caption{Performance of TAP-SVRG and DAP-SVRG for nuclear regularized objective. We evaluate the convergence of sub-optimality $P(x) - P(x^*)$ and $\|x - x^*\|_2^2$  in terms of both epoch and time.} 
	\label{comps}
\end{figure}

In this section, compare our method with DAP-SGD \cite{li2016make}. We compare our DAP-SVRG method with TAP-SVRG, DAP-SGD constant and DAP-SGD decay. DAP-SGD constant denotes decoupled asynchronous proximal stochastic gradient descent method with constant learning rate, DAP-SGD decay denote decoupled asynchronous proximal stochastic gradient descent method with decreasing learning rate, where $\eta_s = \frac{\eta}{(s+1)^\beta} $. In our experiment, learning rate $\eta$ is tuned from $\{10^{-2}, 10^{-3}, 10^{-4} \}$, and $\beta$ is tuned from $\{0.1,0.3,0.5,0.7,0.9 \}$.  All experimental results are shown in Figure \ref{comps}. At first iterations, DAP-SGD decay is faster than DAP-SVRG. Because of the variance of stochastic gradient, we have to reduce the learning rate for SGD method to guarantee convergence. DAP-SVRG's learning rate keeps constant because the variance of stochastic gradient goes to zero. Thus, DAP-SVRG outperforms DAP-SGD decay after several epochs. It is clear that DAP-SVRG converges faster than DAP-SGD decay in terms of both epoch and time. 

\section{Conclusion}
In this paper, we propose a decoupled asynchronous proximal stochastic variance reduced gradient descent method (DAP-SVRG). We use variance reduction technique, and accelerate the convergence rate of traditional asynchronous proximal stochastic gradient method. Additionally, by making workers do proximal mapping step, we relieve server from time-consuming operations. We further provide theoretical analysis of convergence property for DAP-SVRG, and it is proved to linear convergence when learning rate is constant. Experimental results also demonstrate our analysis.

\bibliographystyle{plain}
\bibliography{DAProx-SVRG}  % sigproc.bib is the name of the Bibliography in this case
\clearpage
\appendix

\section{Proof of Lemma \ref{lem1}}

\begin{proof}[Lemma \ref{lem1}]

Because the definition of $u_t$ and $v_t$, we know that $\mathbb{E} u_t = \nabla f(x_t)$ and $\mathbb{E} v_{t} = \nabla f(x_{d(t)})$, then we have: 
\begin{eqnarray}
\mathbb{E} \|v_{t} - \nabla f(x_{d(t)})\|_2^2 & = & \mathbb{E} \| v_{t} - \nabla f(x_{d(t)}) - u_{t} + u_{t} + \nabla f(x_t) - \nabla f(x_t) \|_2^2 \nonumber \\ 
&\leq & 2\mathbb{E} \|v_t - u_t  - (\nabla f(x_d(t))- \nabla f(x_t)) \|_2^2
+ 2\mathbb{E} \|u_t - \nabla f(x_t)\|_2^2 \nonumber \\
&\leq & 2\mathbb{E} \|\nabla f_i(x_{d(t)}) - \nabla f_i(x_t) \|_2^2 + 2\mathbb{E} \|u_t - \nabla f(x_t)\|_2^2 \nonumber \\
&\leq & 2 L^2 \mathbb{E} \|x_{d(t)} - x_t \|_2^2 + 2\mathbb{E} \|u_t - \nabla f(x_t)\|_2^2 \nonumber \\
&\leq & 2 L^2 \tau \sum\limits_{k=d(t)}^{t-1}  \mathbb{E} \|x_{k+1} - x_k \|_2^2 + 2 \mathbb{E} \| u_t - \nabla f(x_t) \|_2^2
\end{eqnarray}
where the first inequality follows from the Lemma 3 in \cite{reddi2015variance}, the second inequality follows from this equation $\mathbb{E} \|\xi  - \mathbb{E} \xi\|_2^2 = \mathbb{E}\|\xi\|_2^2 - \|\mathbb{E}\xi\|_2^2$, the third inequality follows from that $\nabla f_i(x) $ is Lipschitz continuous and the last inequality follows from the assumption that the maximum time delay has an upper bound $\tau$. 

As per \cite{li2016make}, the update of $x_{d(t)}'$ in the worker can be represented as:
\begin{eqnarray}
x_{d(t)}' &=& \text{Prox}_{\eta,h} (x_{d_t} - \eta v_t) \nonumber \\
&=& \arg \min_{y} \frac{1}{2\eta} \|y - (x_{d(t)} - \eta v_t) \|_2^2 + h(y)
\end{eqnarray}
we define $\partial h(x) $ as the subgradient of $h(x)$, then we have:
\begin{eqnarray}
\frac{1}{\eta} (x_{d(t)} - x_{d(t)}') - v_{t} \in \partial h(x_{d(t)}')
\end{eqnarray}
then the following inequality holds that:
\begin{eqnarray}
\label{l1q2}
\mathbb{E} \|v_{t} - \nabla f(x_{d(t)})\|_2^2  &\leq& 2 L^2  \eta^2 \tau  \sum\limits_{k=d(t)}^{t-1}  \mathbb{E}\|v_k + \partial h(x_{d(k)}') \|_2^2 + 2\mathbb{E} \|u_t - \nabla f(x_t)\|_2^2 \nonumber \\ 
&=& 2 L^2  \eta^2 \tau \sum\limits_{k=d(t)}^{t-1}  \mathbb{E} \|v_k + \partial h(x_{d(k)}') - \nabla f(x_{d(k)}) + \nabla f(x_{d(k)}) \|_2^2 \nonumber \\
&  &+ 2\mathbb{E} \|u_t - \nabla f(x_t)\|_2^2 \nonumber \\  
&\leq &4L^2\eta^2 \tau \sum\limits_{k=d(t)}^{t-1} \mathbb{E}\|\partial h(x_{d(k)}') + \nabla f(x_{d(k)}) \|_2^2  \nonumber \\
&&+ 4L^2\eta^2 \tau  \sum\limits_{k=d(t)}^{t-1} \mathbb{E} \|v_k - \nabla f(x_{d(k)}) \|_2^2  + 2\mathbb{E} \|u_t - \nabla f(x_t)\|_2^2
\end{eqnarray}

Follow the convergence proof in \cite{ee236c6}, we define $x_{d(t)}' = x_{d(t)} - \eta G$, then we have $G = v_t + \partial h(x_{d(t)}')$.  Because $\nabla f(x)$ is Lipschitz continuous, then:
\begin{eqnarray}
\label{l1q1}
f(x_{d(t)}' ) & \leq& f(x_{d(t)}) -  \eta \nabla \left< f(x_{d(t)}), G \right> + \frac{L\eta^2}{2} \|G\|_2^2  \nonumber \\
&\leq & f(x_{d(t)}) -  \eta \nabla \left< f(x_{d(t)}), G \right> + \frac{\eta}{2} \|G\|_2^2 
\end{eqnarray}  
where we set $\eta \leq \frac{1}{L}$. Then $P(x_{d(t)}')$ can be bounded in the following manner:
\begin{eqnarray}
P(x_{d(t)}') &=& f(x_{d(t)}' ) + h(x_{d(t)}') \nonumber \\
&\leq & f(x_{d(t)}) - \eta  \left< \nabla f(x_{d(t)}), G \right> + \frac{\eta}{2} \|G\|_2^2 +  h(x_{d(t)}') \nonumber \\ 
&\leq & f(z) - \left< \nabla f(x_{d(t)}), z-x_{d(t)}\right> - \frac{\mu}{2}\|z-x_{d(t)}\|_2^2 -  \eta  \left< \nabla f(x_{d(t)}), G \right>  \nonumber \\ 
&&+ \frac{\eta}{2} \|G\|_2^2 +  h(z) -\left< \partial h(x_{d(t)}'), z-x_{d(t)}' \right>  \nonumber \\
& = & f(z) - \left< \nabla f(x_{d(t)}), z-x_{d(t)}\right> - \frac{\mu}{2}\|z-x_{d(t)}\|_2^2 -  \eta  \left< \nabla f(x_{d(t)}), G \right>  \nonumber \\ 
&&+ \frac{\eta}{2} \|G\|_2^2 +  h(z) -\left< G - v_t, z-x_{d(t)}+ \eta G \right>  \nonumber \\ 
&=& P(z) - \left< \nabla f(x_{d(t)}), z-x_{d(t)}\right> - \frac{\mu}{2}\|z-x_{d(t)}\|_2^2 -  \eta  \left< \nabla f(x_{d(t)}), G \right>  \nonumber \\ 
&& - \frac{\eta}{2} \|G\|_2^2 - \left<G, z-x_{d(t)} \right> + \left<v_t, z-x_{d(t)} \right> + \eta \left< v_t, G \right>
\end{eqnarray} 
where the first inequality comes form inequality (\ref{l1q1}), the second inequality comes from that both $f(x)$ and $h(x)$ are convex functions. We let $z=x_{d(t)}$, then the following inequality holds:
\begin{eqnarray}
\label{l1q3}
P(x_{d(t)}') \leq P(x_{d(t)})  - \frac{\eta}{2} \|G\|_2^2  - \eta \left<G, \nabla f(x_{d(t)}) - v_t \right>
\end{eqnarray}
Thus,  $\mathbb{E}\|\partial h(x_{d(t)}') + \nabla f(x_{d(t)}) \|_2^2$ has an upper bound:
\begin{eqnarray}
\label{l1q4}
 \mathbb{E}\|\partial h(x_{d(t)}') + \nabla f(x_{d(t)}) \|_2^2 &=&  \mathbb{E}\|\partial h(x_{d(t)}') + \nabla f(x_{d(t)}) - v_t+ v_t\|_2^2 \nonumber \\
 & = & \mathbb{E}\|v_t + \partial h(x_{d(t)}')  \|_2^2 + \mathbb{E} \| v_t -\nabla f(x_{d(t)}) \|_2^2 \nonumber \\
 && + 2\mathbb{E}\left<v_t + \partial h(x_{d(t)}'), \nabla f(x_{d(t)}) - v_t   \right> \nonumber \\
 &\leq& \mathbb{E} \| v_t -\nabla f(x_{d(t)}) \|_2^2  + \frac{2}{\eta} \mathbb{E} \left[P(x_{d(t)}) - P(x_{d(t)}') \right] 
\end{eqnarray}
where the inequality follows from (\ref{l1q3}).  Then, substituting (\ref{l1q4}) into (\ref{l1q2}) and adding both sides from $t=sm$ to $t=sm+m-1$, then we have:
\begin{eqnarray}
\sum\limits_{t=sm}^{sm+m-1} \mathbb{E}\| v_t - \nabla f(x_{d(t)})\|_2^2 &\leq& 8 L^2\eta^2 \tau^2 \sum\limits_{t=sm}^{sm+m-1} \mathbb{E} \|v_t - \nabla f(x_{d(t)}) \|_2^2 \nonumber \\
&& + 8 L^2\eta \tau^2 \sum\limits_{t=sm}^{sm+m-1} \mathbb{E} [P(x_t) - P(x_t')] \nonumber \\
&& + 2\sum\limits_{t=sm}^{sm+m-1}  \mathbb{E} \|u_t - \nabla f(x_t)\|_2^2
\end{eqnarray}
Finally, the following inequality holds:
\begin{eqnarray}
\sum\limits_{t=sm}^{sm+m-1} \mathbb{E}\| v_t - \nabla f(x_{d(t)})\|_2^2  &\leq& \frac{8 L^2  \tau^2 \eta}{1-8L^2\eta^2\tau^2} \sum\limits_{t=sm}^{sm+m-1}\mathbb{E} [P(x_t) - P(x_t')] \nonumber \\
&& +  \frac{2}{1-8L^2\eta^2\tau^2} \sum\limits_{t=sm}^{sm+m-1}  \mathbb{E} \|u_t - \nabla f(x_t)\|_2^2
\end{eqnarray}
\end{proof}

\section{Proof of Theorem \ref{thm1}}

\begin{proof}
Because  the update rule  $x_{t+1} - x_t = x_{d(t)}' - x_{d(t)} $, then we have:
\begin{eqnarray}
\label{ineq10}
\mathbb{E} \| x_{t+1} - x^*\|_2^2 &=& \mathbb{E} \|x_t - x^* - x_{d(t)}' + x_{d(t)} \|_2^2 \nonumber \\
&=& \mathbb{E} \| x_t - x^*\|^2_2 + \mathbb{E} \|x_{d(t)}' - x_{d(t)}\|_2^2 + 2\mathbb{E} \left< x_{d(t)}' - x_{d(t)}, x_t - x^* \right> \nonumber \\
&=& \mathbb{E} \| x_t - x^*\|^2_2 + \underbrace{\mathbb{E} \|x_{d(t)}' - x_{d(t)}\|_2^2 + 2\mathbb{E} \left< x_{d(t)}' - x_{d(t)}, x_{d(t)} - x^* \right> }_{Q_1} \nonumber  \\
&& +  \underbrace{2\mathbb{E} \left< x_{d(t)}' - x_{d(t)}, x_t - x_{d(t)} \right> }_{Q_2}
\end{eqnarray}
Because $f(x) $ and $h(x)$ are both convex functions, then it holds that:
\begin{eqnarray}
P(x^* ) & = & f(x^*) + h(x^*) \nonumber \\
& \geq & f(x_{d(t)}) + \left< \nabla f(x_{d(t)}), x^* - x_{d(t)} \right>  + h(x_{d(t)}') + \left<\partial h(x_{d(t)}') , x^* - x_{d(t)}'  \right> \nonumber \\
& \geq & f(x_{d(t)}') - \left< \nabla f(x_{d(t)}), x_{d(t)}' - x_{d(t)} \right> - \frac{L}{2} \| x_{d(t)}' - x_{d(t)}\|_2^2  \nonumber \\
&& + \left< \nabla f(x_{d(t)}), x^* - x_{d(t)} \right>  + h(x_{d(t)}')  + \left< \frac{1}{\eta} (x_{d(t)} - x_{d(t)}') - v_t  , x^* - x_{d(t)}'  \right> \nonumber \\
&\geq & P(x_{d(t)}') + \left< \nabla f(x_{d(t)}) - v_t , x^*- x_{d(t)}' \right> + \frac{1}{\eta} \left< x_{d(t)} - x_{d(t)}, x^* - x_{d(t)} \right> \nonumber \\
&& + \frac{1}{2\eta} \|x_{d(t)} - x_{d(t)}'\|_2^2
\end{eqnarray} 
where the second inequality follows from that $\nabla f(x)$ is Lipschitz continuous, and the last inequality holds by setting $\eta \leq \frac{1}{L}$. Then take expectation on both sides, we have:
\begin{eqnarray}
\label{ineq5}
- \mathbb{E}[P(x_{d(t)}')- P(x^*)] + \underbrace{ \mathbb{E} \left< v_t - \nabla f(x_{d(t)}), x^*-x_{d(t)}' \right>}_{Q_3} \geq \nonumber \\
 + \frac{1}{\eta} \mathbb{E}\left< x_{d(t)} - x_{d(t)}' , x^* - x_{d(t)} \right> + \frac{1}{2\eta} \mathbb{E}\| x_{d(t)} - x_{d(t)}'\|_2^2 
\end{eqnarray} 

Define 
$y = Prox_{\eta, h} \left( x_{d(t)} - \eta  \nabla f(x_{d(t)})\right) $, and it is independent of sample $i$. then we have:
\begin{eqnarray}
Q_3 &=& \mathbb{E} \left<  v_t - \nabla f(x_{d(t)}), x^*- y \right> +  \mathbb{E} \left<  v_t - \nabla f(x_{d(t)}), y- x_{d(t)}' \right> \nonumber \\
&=& \mathbb{E} \left< v_t - \nabla f(x_{d(t)}), y- x_{d(t)}' \right>  \nonumber \\
&\leq & \mathbb{E}\|v_t -\nabla f(x_{d(t)})\|_2    \|y-x_{d(t)}'\|_2\nonumber \\
&\leq & \eta  \mathbb{E} \|v_t - \nabla f(x_{d(t)})\|_2^2
\end{eqnarray}
where the second equality follows from that $\mathbb{E} v_t = \nabla f(x_{d(t)})$, the first inequality follows from the Cauchy-Schwarz inequality, and the last inequality follows from the non-expansive property of proximal operator. Therefore, $Q_1$ is upper bounded as follows:
\begin{eqnarray}
Q_1 &\leq& - 2\eta \mathbb{E}[P(x_{d(t)}')- P(x^*)] + 2\eta^2  \mathbb{E} \|v_t - \nabla f(x_{d(t)})\|_2^2
\end{eqnarray}
It is easy to know the upper bound of $Q_2$:
\begin{eqnarray}
Q_2  &=& 2\mathbb{E} \left< x_{d(t)}' - x_{d(t)}, x_t - x_{d(t)} \right> \nonumber \\
&\leq & \mathbb{E} \|x_{d(t)}' - x_{d(t)} \|_2^2 + \mathbb{E} \|x_t - x_{d(t)} \|_2^2 \nonumber \\
& = & \mathbb{E} \|x_{d(t)}' - x_{d(t)} \|_2^2 + \mathbb{E} \| \sum\limits_{k=d(t)}^{t-1} ( x_{k+1} - x_k )\|_2^2 \nonumber \\
&\leq & \eta^2 \mathbb{E} \|v_t + \partial h(x_{d(t)}')  \|_2^2 + \tau \eta^2 \sum \limits_{k=d(t)}^{t-1} \mathbb{E} \|v_k + \partial h(x_{d(k)}') \|_2^2 \nonumber \\ 
&\leq & 2\eta^2 \left( \mathbb{E} \|v_t - \nabla f(x_{d(t)})  \|_2^2  + \mathbb{E} \| \nabla f(x_{d(t)}) + \partial h(x_{d(t)}')  \|_2^2  \right)  \nonumber \\
& & +   2\eta^2 \tau \sum\limits_{k=d(t)}^{t-1} \left( \mathbb{E} \|v_k - \nabla f(x_{d(k)})  \|_2^2  + \mathbb{E} \| \nabla f(x_{d(k)}) + \partial h(x_{d(k)}')  \|_2^2  \right) \nonumber \\
&\leq & 2\eta^2 \left( 2\mathbb{E} \|v_t - \nabla f(x_{d(t)})  \|_2^2  + \frac{2}{\eta} \mathbb{E} \left[P(x_{d(t)}) - P(x_{d(t)}')\right]  \right)  \nonumber \\
&& +  2\eta^2 \tau \sum\limits_{k=d(t)}^{t-1} \left( 2\mathbb{E} \|v_k - \nabla f(x_{d(k)})  \|_2^2  + \frac{2}{\eta} \mathbb{E} \left[P(x_{d(k)}) - P(x_{d(k)}')\right]  \right) 
\end{eqnarray} 
where the second and last inequality use the assumption that maximum time delay is $\tau$, the last inequality follows from inequality (\ref{l1q4}).

Substituting $Q_1$ and $Q_2$ into inequality (\ref{ineq10}), and 
adding inequality (\ref{ineq10}) from $t=sm$ to $t=sm+m-1$, the following inequality holds:
\begin{eqnarray}
\mathbb{E} \|x_{sm+m-1} - x^*\|_2^2 &\leq &\mathbb{E}  \|\tilde{x}_s - x^*\|_2^2 - 2\eta \sum\limits_{t=sm}^{sm+m-1} \mathbb{E}[P(x_t)- P(x^*)] \nonumber \\
&& + (6\eta^2 + 4\eta^2 \tau^2 ) \sum\limits_{t=sm}^{sm+m-1} \mathbb{E} \|v_t - \nabla f(x_{d(t)}) \|_2^2 \nonumber \\
&&  +  (6\eta + 4\eta \tau^2) \sum\limits_{t=sm}^{sm+m-1} \mathbb{E} \left[ P(x_t) - P(x_t') \right]
\end{eqnarray}
where we use $\tilde{x}_s=x_{sm}$. Then assuming there is a very small positive constant $\varepsilon$ that satisfies $P(x_t) - P(x_t') \leq \varepsilon (P(x_t) - P(x^*))$ and according to Lemma \ref{lem1}, we have:
\begin{eqnarray}
\mathbb{E} \|x_{sm+m-1} - x^*\|_2^2 &\leq &\mathbb{E}  \|\tilde{x}_s - x^*\|_2^2 - 2\eta \sum\limits_{t=sm}^{sm+m-1} \mathbb{E}[P(x_t)- P(x^*)] \nonumber \\
&& + \frac{8L^2 \eta^2 \tau \varepsilon (6\eta^2 + 4 \eta^2 \tau^2)}{1-8L^2\eta^2\tau^2}\sum\limits_{t=sm}^{sm+m-1} \left[ P(x_t) - P(x^*) \right] \nonumber \\
&& + \frac{2(6\eta^2 + 4 \eta^2 \tau^2)}{1-8L^2\eta^2\tau^2}\sum\limits_{t=sm}^{sm+m-1}\mathbb{E} \|u_t - \nabla f(x_{d(t)}) \|_2^2  \nonumber \\
&& +  (6\eta + 4\eta \tau^2)\varepsilon \sum\limits_{t=sm}^{sm+m-1} \mathbb{E} \left[ P(x_t) - P(x^*) \right]
\end{eqnarray}
As per \cite{xiao2014proximal}, we know 
\begin{eqnarray}
 \|\tilde{x}_s - x^* \|_2^2 \leq \frac{2}{\mu} \left[ P(\tilde{x}_s) -P(x^*) \right]  \\
 \mathbb{E} \|u_t - \nabla f(x_t)\|_2^2 \leq 4L \left[ P(x_t) - P(x^*) + P(\tilde{x}) - P(x^*) \right]
\end{eqnarray}
and 
\begin{eqnarray}
\sum\limits_{t=sm}^{sm+m-1} [P(x_t) - P(x^*)] \geq m [P(\tilde x_{s+1}) - P(x^*) ]
\end{eqnarray}
finally we have:
\begin{eqnarray}
\left( 2\eta - \frac{8L^2 \eta^2 \tau \varepsilon (6\eta^2 + 4 \eta^2 \tau^2)}{1-8L^2\eta^2\tau^2} -  (6\eta + 4\eta \tau^2)\varepsilon - \frac{8L(6\eta^2 + 4 \eta^2 \tau^2)}{1-8L^2\eta^2\tau^2} \right)m \mathbb{E} \left[ P(\tilde{x}_{s+1}) - P(x^*) \right] \nonumber \\
\leq \left(\frac{2}{\mu} + \frac{8L(6\eta^2 + 4 \eta^2 \tau^2)m}{1-8L^2\eta^2\tau^2} \right) \mathbb{E} \left[ P(\tilde{x}_s) - P(x^*)\right]
\end{eqnarray}

\end{proof}

\end{document}